\documentclass[letterpaper]{article} 
\usepackage{aaai2026}  
\usepackage{times}  
\usepackage{helvet}  
\usepackage{courier}  
\usepackage[hyphens]{url}  
\usepackage{graphicx} 
\usepackage{natbib}  
\usepackage{caption} 
\usepackage{algorithm}
\usepackage{algorithmic}

\usepackage{dsfont}
\usepackage{booktabs} 
\usepackage{amsmath}
\usepackage{multirow}
\usepackage{pifont}
\usepackage{xcolor}
\newcommand{\cmark}{\textcolor{black!60!black}{\ding{51}}}  
\newcommand{\xmark}{\textcolor{black}{\ding{55}}}             

\usepackage{newfloat}
\usepackage{listings}
\DeclareCaptionStyle{ruled}{labelfont=normalfont,labelsep=colon,strut=off} 
\floatstyle{ruled}
\newfloat{listing}{tb}{lst}{}
\floatname{listing}{Listing}
\title{Easy to Learn, Yet Hard to Forget: Towards Robust Unlearning Under Bias}
\author{
    JuneHyoung Kwon\equalcontrib \textsuperscript{\rm 1}, 
    MiHyeon Kim\equalcontrib \textsuperscript{\rm 3}, 
    Eunju Lee\textsuperscript{\rm 2}, \\ 
    Yoonji Lee\textsuperscript{\rm 1}, 
    Seunghoon Lee\textsuperscript{\rm 2}, 
    YoungBin Kim\textsuperscript{\rm 1, 2}
}
\affiliations{
    \textsuperscript{\rm 1} Department of Artificial Intelligence, Chung-Ang University\\
    \textsuperscript{\rm 2} Graduate School of Advanced Imaging Sciences, Multimedia and Film, Chung-Ang University\\
    \textsuperscript{\rm 3} KT Corporation\\
    dirchdmltnv@cau.ac.kr, mihyeon.gim@kt.com, dmswn5829@cau.ac.kr, \\
    pioneer0305@cau.ac.kr, remomare@cau.ac.kr, ybkim85@cau.ac.kr
}

\begin{document}
\pagestyle{plain}
\thispagestyle{plain}

\maketitle

\begin{abstract}
Machine unlearning, which enables a model to forget specific data, is crucial for ensuring data privacy and model reliability. However, its effectiveness can be severely undermined in real-world scenarios where models learn unintended biases from spurious correlations within the data. This paper investigates the unique challenges of unlearning from such biased models. We identify a novel phenomenon we term ``shortcut unlearning," where models exhibit an ``easy to learn, yet hard to forget" tendency. Specifically, models struggle to forget easily-learned, bias-aligned samples; instead of forgetting the class attribute, they unlearn the bias attribute, which can paradoxically improve accuracy on the class intended to be forgotten. To address this, we propose CUPID, a new unlearning framework inspired by the observation that samples with different biases exhibit distinct loss landscape sharpness. Our method first partitions the forget set into causal- and bias-approximated subsets based on sample sharpness, then disentangles model parameters into causal and bias pathways, and finally performs a targeted update by routing refined causal and bias gradients to their respective pathways. Extensive experiments on biased datasets including Waterbirds, BAR, and Biased NICO++ demonstrate that our method achieves state-of-the-art forgetting performance and effectively mitigates the shortcut unlearning problem.
\end{abstract}


\section{Introduction}

Machine unlearning, the task of efficiently removing specific data's influence from a pre-trained model, has become critical for trustworthy AI ~\cite{shaik2024exploring}. This capability is driven by needs ranging from data protection regulations like the `right to be forgotten'~\cite{hoofnagle2019european} to curating models by removing harmful or biased content. Existing unlearning research addresses the removal of various types of information, such as specific data instances~\cite{cha2024learning}, entire classes~\cite{zhou2025decoupled}, or abstract concepts~\cite{wu2025unlearning} learned by the model. These varied approaches are often unified by a critical, implicit assumption that the target information is cleanly separable within the model's parameters.

However, this assumption of separability rarely holds under the biased conditions common in real-world scenarios. Deep neural networks are often trained on inherently biased datasets, where spurious attributes (e.g., a ``water" background) are highly correlated with class labels (e.g., ``waterbird") ~\cite{nam2020learning, lim2023biasadv}. These models are particularly adept at exploiting these correlations, learning ``shortcuts" that are simpler than the true causal features ~\cite{geirhos2020shortcut, kwon2024learning}. This leads to highly entangled representations where the features defining the class are intertwined with those defining the bias ~\cite{lee2021learning}. While the detrimental effect of such shortcuts on model generalization is well-documented, their impact on machine unlearning has remained largely unexplored. How can a model forget a ``waterbird" if its understanding of the class is fundamentally tied to the ``water" background? This confounding of features poses a significant, unaddressed challenge to existing unlearning paradigms.

To this end, we conduct the first investigation into the behavior of unlearning algorithms under data bias, revealing two striking phenomena. First, we observe an ``easy to learn, yet hard to forget" asymmetry: models struggle to unlearn
bias-aligned samples (i.e., those where a spurious feature correctly predicts the class ), yet readily forget bias-conflicting samples (those where the shortcut is misleading ). Second, we identify a paradoxical debiasing effect, where the unlearning process unexpectedly improves accuracy on the bias-conflicting samples of the very class intended for forgetting. Taken together, these findings expose a critical failure mode which we term \textbf{shortcut unlearning}. We define this as the phenomenon where a model, when instructed to forget a target class, primarily erases its reliance on spurious shortcut features rather than the actual causal features of the class itself. This indicates that existing methods fail to perform their intended function, instead merely altering the model's reliance on spurious correlations.

The central challenge, therefore, is to enforce the forgetting of causal, class-related information, even when the model's internal representations are steering the process towards forgetting the bias instead. A robust unlearning algorithm must be able to intervene surgically, rather than applying a uniform update. Such a strategy necessitates two key capabilities: first, a method to identify and disentangle the distinct model parameters that rely on causal versus shortcut features, and second, a mechanism to apply tailored updates to each parameter subset accordingly.

To effectively forget a target class under biased conditions, we propose \textbf{Causal Unlearning via Pathway Identification and Disentanglement (CUPID)}. Our framework is motivated by principles from loss landscape analysis ~\cite{foret2020sharpness}, which connect the sharpness of the loss function to model behavior. We posit that the entanglement between causal and shortcut features manifests as distinct sharpness signatures, which can be leveraged to tackle the shortcut unlearning problem through a three-stage process. First, Sharpness-Aware Partitioning slices the forget set based on local loss sharpness. This stage is designed to isolate samples primarily classified via causal features from those reliant on shortcuts, which is essential for identifying the distinct neural pathways activated by each type. Second, Causal Pathway Identification identifies the causal pathway by selecting a subset of parameters exhibiting high curvature and large magnitude. The purpose of this step is to precisely target the parameters for updates, thereby preventing the model from unintentionally exploiting the bias during unlearning. Finally, our Targeted Pathway Update mechanism applies tailored gradients to each pathway. This allows for the precise erasure of class information in the causal pathway while appropriately managing the bias pathway, ensuring a robust and targeted unlearning outcome.

Experiments on standard biased datasets—including Waterbirds  ~\cite{sagawa2019distributionally}, BAR ~\cite{nam2020learning}, and Biased NICO++ ~\cite{zhang2023nico++}—validate the superiority of our proposed method. CUPID significantly outperforms existing unlearning methods by achieving the lowest forget accuracy, which demonstrates its ability to effectively erase a target class under biased conditions. Furthermore, its strong and balanced performance on both bias-aligned and bias-conflicting samples provides direct evidence that CUPID effectively mitigates the shortcut unlearning problem. Our contributions are

\begin{itemize}
    \item We identify and formalize shortcut unlearning, a critical failure mode of unlearning algorithms in the presence of data bias.
    \item We propose \textbf{Causal Unlearning via Pathway Identification and Disentanglement (CUPID)}, a novel framework that leverages loss landscape geometry to disentangle and selectively update causal and bias gradients.
    \item We provide a comprehensive empirical validation demonstrating that CUPID achieves superior unlearning performance on biased datasets by selectively forgetting causal information.
\end{itemize}

\section{Related Work}

\subsection{Machine Unlearning}
The foundational approach to machine unlearning, often termed exact unlearning, involves retraining a model from scratch~\cite{thudi2022unrolling, shaik2024exploring}. This serves as a theoretical gold standard, but its prohibitive computational cost often renders the approach impractical ~\cite{dwork2006our}. To address this limitation, research has shifted towards developing approximate unlearning methods that aim to efficiently simulate the result of retraining without the full computational burden.

Approximate unlearning methods can be broadly categorized by their underlying mechanisms. One line of work aims to directly degrade the model’s performance on the forget set by maximizing the loss, which is achieved either by increasing the loss on the forget set  ~\cite{thudi2022unrolling} or by injecting adversarial noise to reverse the effects of learning ~\cite{tarun2023fast}. Another category leverages relabeling, which functions as a form of targeted data poisoning. These methods finetune the model on the forget set using randomly assigned labels ~\cite{golatkar2020eternal}, or more strategically, reassign forget samples to their nearest incorrect classes or an auxiliary shadow class to shift or dissolve decision boundaries ~\cite{chen2023boundary}. A third group of approaches employs knowledge distillation, training the unlearned model to diverge from the original model on the forget set  ~\cite{zhou2025decoupled}, while optionally encouraging alignment on a retain set to preserve model utility ~\cite{kurmanji2023towards, chundawat2023can, kim2024layer}. Lastly, rather than modifying the entire parameter space, some methods restrict updates to a subset of layers ~\cite{goel2022towards} or selectively identify and alter a sparse set of parameters most critical to the forget set  ~\cite{jia2023model, fan2023salun, cha2024learning}, thereby minimizing collateral forgetting on unrelated data.

However, existing methods face two key challenges: they largely assume target information is cleanly separable, a premise that fails when the target class is highly correlated with task-irrelevant features ~\cite{choi2025distribution, zhao2024makes}; and they often rely on access to the retain set, a requirement that is frequently infeasible due to privacy or data storage constraints  ~\cite{li2025machine}. To this end, we propose a method designed for the challenging condition where a target class is highly correlated with spurious shortcuts, which also, critically, operates without requiring access to the retain set.

\begin{figure*}[t]
\centering
\includegraphics[width=1.0\textwidth]{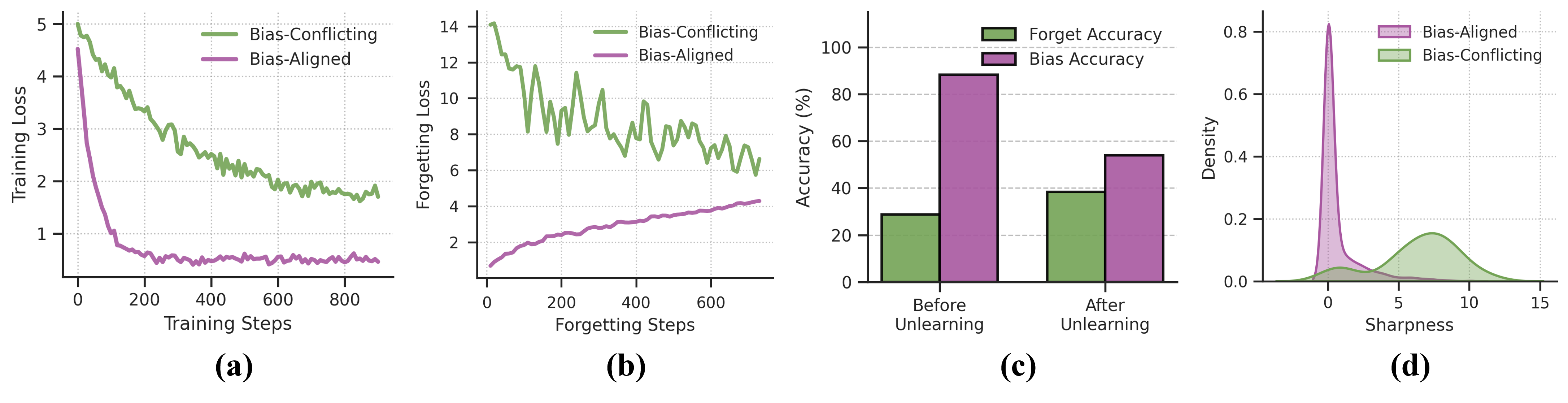}
\caption{{Analysis of Shortcut Unlearning.} (a) Rapid learning of bias-aligned samples. (b) Slow forgetting of bias-aligned samples. (c) Linear probing accuracy confirming shortcut removal. (d) Sharpness distributions distinguishing sample types.}
\label{fig:analysis}
\end{figure*}

\section{Preliminary}

Let $\mathcal{D}_{\text{train}}=\left\{(x_i, y_i) \right\}_{i=1}^N$ be the full training dataset, where $y_i \in \mathcal{Y} = \{1, \dots, K\}$ is the class label, and $f(\cdot; \theta_o)$ is an original model trained on this data. Our work addresses \textbf{class-wise forgetting}, where the goal is to remove the influence of a specific target class, $c_f \in \mathcal{Y}$. The forget set is thus defined as all samples belonging to this class, $\mathcal{D}_f = \{(x, y) \in \mathcal{D}_{\text{train}} \mid y = c_f\}$, and the unlearning process aims to produce an unlearned model $f(\cdot;\theta_u)$ from the original $\theta_o$. Specifically, we tackle this problem under the challenging condition of data bias, where each sample $x_i$ possesses a spurious attribute $b_i$ that is highly correlated with its class label $y_i$. This leads the model to learn a ``shortcut," prompting us to categorize samples within $\mathcal{D}_f$ based on this bias: bias-aligned samples ($\mathcal{D}_f^{\text{BA}}$) are those where the shortcut ($b_i$) aligns with the class ($y_i=c_f$), while bias-conflicting samples ($\mathcal{D}_f^{\text{BC}}$) are those where it fails ~\cite{nam2020learning}. Our method is built upon this fundamental distinction.

\section{Analysis: Unlearning on Biased Models}

To understand the underlying challenges of unlearning on biased models, we analyze how a standard unlearning algorithm interacts with a model $\theta_o$ trained on a biased dataset $\mathcal{D}_{\text{train}}$. As a preliminary step, we confirm that during initial training, the model first learns the ``easy" patterns from bias-aligned samples ($\mathcal{D}^{\text{BA}}$) before fitting to the ``harder" bias-conflicting samples ($\mathcal{D}^{\text{BC}}$), which aligns with findings in prior work ~\cite{nam2020learning} (Figure~\ref{fig:analysis} (a)).

Next, we apply a standard unlearning algorithm, NegGrad ~\cite{thudi2022unrolling}, and track the forgetting loss on the forget set $\mathcal{D}_f$, separating the two sample types. The results, illustrated in Figure~\ref{fig:analysis} (b), reveal a striking asymmetry. The forgetting loss for bias-conflicting samples achieves a significantly higher magnitude than that of bias-aligned samples, indicating that the former are easily forgotten, while the latter are rather difficult to forget. This leads to our first key finding: Easy to learn, yet hard to forget. The model finds it difficult to forget the very samples it learned most easily because its representation has become heavily reliant on the simple, robust shortcut attribute.

Furthermore, a closer look at the loss dynamics reveals a more critical issue: the loss for bias-conflicting samples sharply \textit{decreases}, resulting in a paradoxical debiasing effect where the model improves its predictions for these samples. We conjecture that this occurs because the unlearning process does not target the intended causal features, but rather erases the dominant shortcut representation the model relied upon. Essentially, when tasked to forget, the model takes the ``easy way out" by forgetting the most prominent pattern it learned---the shortcut.

Our subsequent experiments validate this hypothesis. We observe that the accuracy on the bias-conflicting subset of the forget set indeed paradoxically increases after unlearning. A linear probe experiment provides more direct evidence: when we attach a simple linear classifier to the model's frozen representations to predict the bias label $b_i$, its accuracy drops significantly on the unlearned model $\theta_u$ compared to the original $\theta_o$ as shown in Figure~\ref{fig:analysis} (c). 
This confirms that the unlearning process primarily targeted and removed the shortcut representation, providing a detailed characterization of the \textbf{shortcut unlearning} phenomenon.

\section{Method}

\begin{figure*}[t]
\centering
\includegraphics[width=0.924\textwidth]{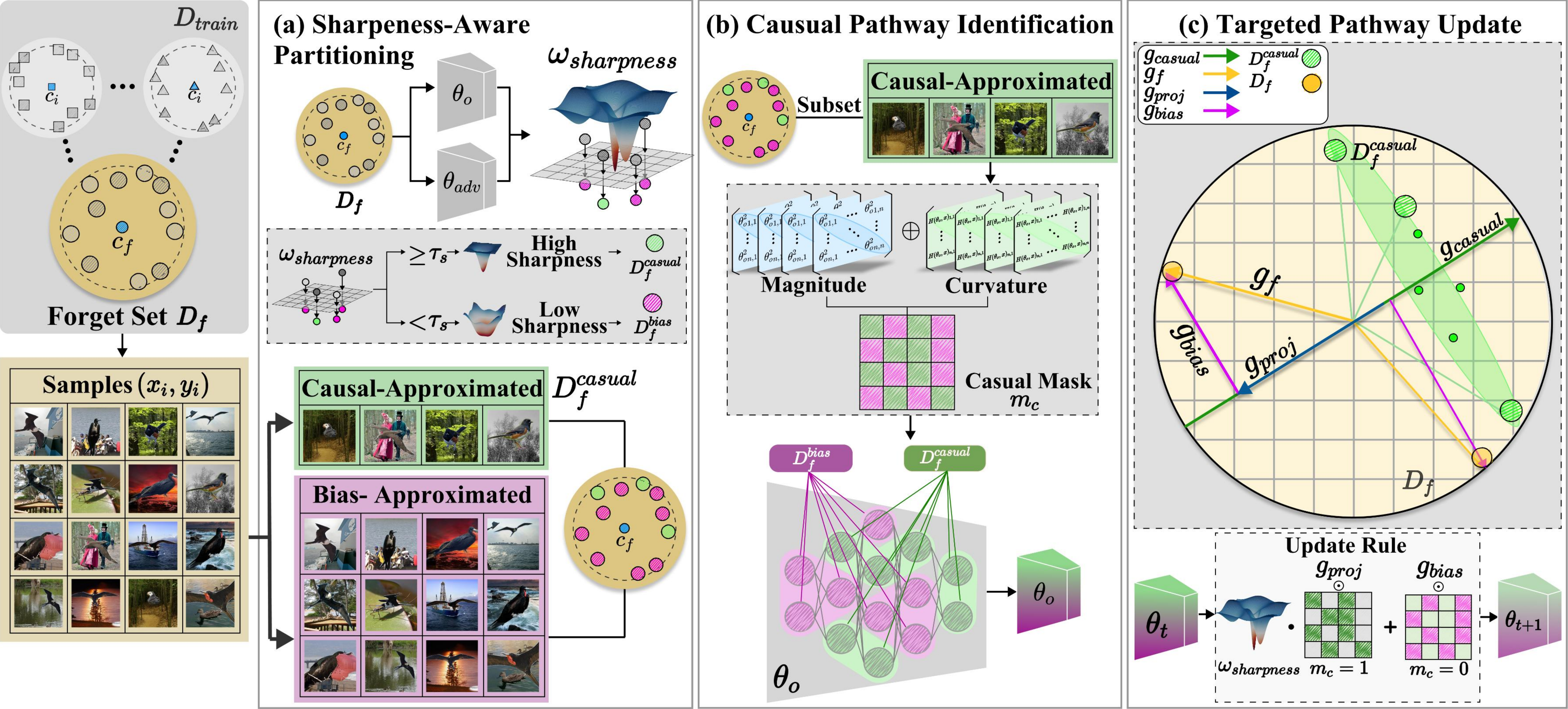}
\caption{{The CUPID Framework.} Our proposed method consists of three stages. (a) Sharpness-Aware Partitioning divides the forget set into causal- and bias-approximated subsets based on local loss sharpness. (b) Causal Pathway Identification disentangles the model's parameters into a causal pathway and a bias pathway. (c) The Targeted Pathway Update applies distinct gradients to each pathway to perform a surgical unlearning.}
\label{fig:overview}
\end{figure*}

Our analysis reveals that standard unlearning algorithms fail due to shortcut unlearning. This failure stems from a critical blindness: their inability to distinguish between causal and shortcut features, which are learned from different sample types. A uniform update inevitably targets the most dominant patterns---the shortcuts. To perform a more surgical intervention, we turn to the geometry of the loss landscape, where we argue the necessary signal to differentiate these features is encoded.

Our key insight is that the distinct geometric properties of the loss landscape for bias-aligned and bias-conflicting samples can be explained through the lens of generalization. Modern deep learning has established a strong connection between a model's generalization capability and the flatness of the loss minima it converges to; models that generalize well tend to find wide, flat minima~\cite{foret2020sharpness, wen2022does, chen2023does}. In a biased dataset, the model primarily learns to generalize based on the dominant, ``easy-to-learn" shortcut features present in the majority of bias-aligned samples~\cite{springer2024sharpness}. Consequently, the region of the loss landscape corresponding to these ``easy," generalized samples is expected to be flat, exhibiting low curvature~\cite{zhou2023imbsam}. Conversely, the ``hard" bias-conflicting samples, which the model struggles to fit, are expected to occupy sharp, high-curvature regions~\cite{zhou2023imbsam}. 

We empirically validate this hypothesis by plotting the distribution of local loss sharpness values for both sample types (Figure~\ref{fig:analysis} (d)). The results clearly show that bias-aligned samples cluster in a narrow distribution near zero, confirming their association with flat regions. In contrast, bias-conflicting samples exhibit a much broader distribution with a long tail into higher sharpness values, confirming their location in sharper, more sensitive areas of the landscape.

This empirically verified difference in loss landscape sharpness, rooted in the principles of generalization, offers a principled way to differentiate samples based on how the model processes them. By measuring this local sharpness, we can effectively ``slice" the data and identify the neural pathways associated with each group. To this end, we propose Causal Unlearning via Pathway Identification and Disentanglement (CUPID), a three-stage framework designed to operationalize this insight. Our method begins with Sharpness-Aware Partitioning to separate the forget set, followed by Causal Pathway Identification to isolate the relevant neural circuits, and concludes with a Targeted Pathway Update to perform a surgical unlearning. The overall framework is depicted in Figure~\ref{fig:overview}.

\subsection{Sharpness-Aware Partitioning}
The first stage of our method is to partition the forget set $D_f$ into subsets that approximate the true bias-aligned ($D_f^{BA}$) and bias-conflicting ($D_f^{BC}$) distributions. Our goal is to derive a partition based on the model's own behavior, separating samples that are likely learned via shortcuts from those learned via causal features. 
Building on our analysis above, our approach leverages the insight that ``easy" bias-aligned and ``hard" bias-conflicting samples correspond to flat and sharp regions of the loss landscape, respectively. To quantify this local sharpness for each sample $(x_i,y_i)$ in $D_f$, we measure the sensitivity of the loss to a worst-case perturbation in a small neighborhood around the current parameters. This perspective, which connects local geometry to model behavior, allows us to directly estimate the sharpness for each sample.

We first compute the loss $L(\theta_o, x_i)$ and its gradient $\nabla L(\theta_o, x_i)$ using the original model parameters $\theta_o$. We then create a perturbed set of parameters, $\theta_{\text{adv}}$, by taking a small, normalized step in the direction of the gradient: \begin{equation} \label{eq:1} \theta_{\text{adv}} = \theta_o + \eta \frac{\nabla L(\theta_o, x_i)}{||\nabla L(\theta_o, x_i)||} \end{equation} where $\eta$ is a small step size. The local sharpness for sample $x_i$, denoted as $\omega_{\text{sharpness}}(x_i)$, is then defined as the change in loss resulting from this perturbation: 

\begin{equation} 
\label{eq:2} 
\omega_{\text{sharpness}}(x_i) = L(\theta_{\text{adv}}, x_i) - L(\theta_o, x_i) 
\end{equation} 

After computing $\omega_{\text{sharpness}}$ for all samples in $\mathcal{D}_f$, we partition the set based on these values. Specifically, we set a threshold $\tau_s$ corresponding to the top $k\%$ of sharpness values. Samples with a sharpness value below this threshold are assigned to the bias-approximated set ($\mathcal{D}_f^{\text{bias}}$), while those with a value above $\tau_s$ are assigned to the causal-approximated set ($\mathcal{D}_f^{\text{causal}}$). This allows us to effectively separate the samples based on the model's own processing characteristics, paving the way for a more targeted unlearning procedure.

\begin{table*}[t]
\centering
\label{tab:results}
\setlength{\tabcolsep}{1.7pt}
\small
\resizebox{\textwidth}{!}{%
\begin{tabular}{l c ccccc ccccc ccccc}
\toprule
\multirow{2}{*}{Method} & \multirow{2}{*}{\shortstack{w/o\\retain\\set}} & \multicolumn{5}{c}{Waterbirds} & \multicolumn{5}{c}{BAR} & \multicolumn{5}{c}{Biased NICO++} \\
\cmidrule(lr){3-7} \cmidrule(lr){8-12} \cmidrule(lr){13-17}
& & RA $\uparrow$ & FA $\downarrow$ & $\triangle_{gap}$ $\downarrow$ & WGA $\downarrow$ & MIA $\downarrow$ & RA $\uparrow$ & FA $\downarrow$ & $\triangle_{gap}$ $\downarrow$ & WGA $\downarrow$ & MIA $\downarrow$ & RA $\uparrow$ & FA $\downarrow$ & $\triangle_{gap}$ $\downarrow$ & WGA $\downarrow$ & MIA $\downarrow$ \\
\midrule
Retrain & \xmark & 100.00 & 0.00 & 0.00 & 0.00 & 12.34 & 97.64 & 0.00 & 0.00 & 0.00 & 14.92 & 94.68 & 0.00 & 0.00 & 0.00 & 21.17 \\
NegGrad & \cmark & \textbf{99.81} & 34.96 & 29.17 & 36.41 & 18.45 & \textbf{97.00} & 58.59 & 56.53 & 58.87 & 21.39 & 94.24 & 22.33 & 15.09 & 22.41 & 23.71 \\
Random label & \cmark & 99.94 & 18.20 & 17.63 & 19.08 & 17.92 & 94.49 & 24.97 & 21.19 & 25.08 & 17.88 & 95.38 & 28.72 & 28.86 & 28.87 & 24.29 \\
Bad Teaching & \xmark & 98.74 & 88.35 & 68.56 & 91.75 & 20.09 & 94.68 & 36.27 & 35.22 & 36.45 & 18.74 & 97.03 & 53.61 & 48.22 & 53.85 & 26.93 \\
Boundary Shrink & \cmark & 99.87 & 26.47 & 26.33 & 27.78 & 18.38 & 95.93 & 52.37 & 45.48 & 52.60 & 22.51 & 93.89 & 23.44 & 22.32 & 23.55 & 24.54 \\
Boundary Expand & \cmark & 99.96 & 20.00 & 17.99 & 20.89 & \textbf{18.15} & 94.49 & 30.34 & 29.66 & 30.49 & 18.44 & 95.03 & 29.38 & 29.53 & 29.53 & 24.98 \\
SALUN & \xmark & 99.91 & 37.05 & 20.69 & 38.08 & 18.94 & 95.31 & 38.81 & 39.01 & 39.01 & 18.66 & \textbf{96.42} & 54.27 & 50.97 & 54.53 & 27.37 \\
DELETE & \cmark & 99.91 & 18.42 & 17.85 & 19.30 & 18.32 & 95.93 & 34.86 & 33.80 & 35.03 & 21.26 & 95.72 & 27.84 & 19.82 & 27.94 & 23.56 \\
\textbf{CUPID (Ours)} & \cmark & 99.94 & \textbf{6.91} & \textbf{7.27} & \textbf{7.27} & 18.19 & 95.89 & \textbf{7.70} & \textbf{7.74} & \textbf{7.74} & \textbf{16.77} & 91.80 & \textbf{7.71} & \textbf{7.63} & \textbf{7.75} & \textbf{23.51} \\
\bottomrule
\end{tabular}
}
\caption{{Unlearning from a Heavily Biased Distribution.} We compare various unlearning algorithms on three biased datasets where the ratio of bias-aligned to bias-conflicting samples in the forget class is 99.5:0.5. Our proposed method, CUPID, significantly outperforms all baselines by achieving the lowest FA and the most balanced forgetting (lowest $\triangle_{gap}$ and WGA).}
\label{tab:main}
\end{table*}

\subsection{Causal Pathway Identification}
With the forget set partitioned, the next stage is to identify which parts of the model store the knowledge associated with each data type. This is based on the premise that specific knowledge is often localized within a model's parameters, allowing for targeted modifications ~\cite{patil2023can, menik2023towards}. The goal of this stage is to disentangle the model's parameters, $\theta_o$, into a causal pathway, which is predominantly used for the causal-approximated set ($\mathcal{D}_f^{\text{causal}}$), and a bias pathway, which is predominantly used for the bias-approximated set ($\mathcal{D}_f^{\text{bias}}$).

We characterize a parameter's importance by both its magnitude and its location in a high-curvature region of the loss landscape. This approach is conceptually grounded in foundational work on network simplification, which demonstrated that a parameter's saliency can be effectively estimated by combining its magnitude with the second derivative of the loss function ~\cite{lecun1989optimal}. To identify the parameters crucial for representing causal information, we adapt this principle to define a causal mask, $m_c$. For each parameter $\theta_{o, i}$, we directly compute the mask value as follows:

\begin{equation}
\label{eq:3}
m_c(\theta_{o,i}) = 
\mathds{1} \left( \frac{1}{2} \theta_{o,i}^2 \cdot \mathds{E}_{x \sim \mathcal{D}_f^{\text{causal}}}[H(\theta_o, x)_{ii}] \geq \tau_p \right)
\end{equation}

where $\mathds{1}(\cdot)$ is the indicator function, $H(\theta_o, x)_{ii}$ is the i-th diagonal element of the Hessian matrix, and $\tau_p$ is a hyperparameter set to select a certain percentile (e.g., top 50\%) of the most influential parameters. This mask effectively isolates the causal pathway, allowing us to apply a targeted update in the final stage of our method. The remaining parameters are considered part of the bias pathway.

\subsection{Targeted Pathway Update}

With the forget set partitioned and the neural pathways identified, the final stage is to perform a targeted parameter update. The objective is to precisely erase the causal information encoded in the causal pathway while minimizing unintended alterations to the bias pathway.

To achieve this, we first define a causal gradient direction, $g_{\text{causal}}$, as the average gradient computed over the causal-approximated set, $\mathcal{D}_f^{\text{causal}}$. This vector represents the ideal direction for forgetting the causal features. The overall gradient for the entire forget set, $g_f$ (computed over all of $\mathcal{D}_f$), is then projected onto this causal direction to obtain the component aligned with forgetting causal information, $g_{\text{proj}}$, which is calculated as follows:

\begin{equation}
\label{eq:5}
g_{\text{proj}} = \frac{g_f \cdot g_{\text{causal}}}{||g_{\text{causal}}||^2} g_{\text{causal}}
\end{equation}

The remaining orthogonal component, $g_{\text{bias}} = g_f - g_{\text{proj}}$, is then treated as the gradient primarily related to the shortcut features. The final update rule surgically applies these distinct gradients to their corresponding pathways, as identified by the causal mask $m_c$:

\begin{equation}
\label{eq:6}
\theta_{t+1} \leftarrow \theta_t + \alpha \cdot [ (\omega_{\text{sharpness}} \cdot g_{\text{proj}} \odot m_c) + (g_{\text{bias}} \odot (1 - m_c)) ]
\end{equation}

where $\alpha$ is the learning rate and $\odot$ denotes element-wise multiplication. For the causal pathway (where $m_c=1$), the update is guided by the projected causal gradient, $g_{\text{proj}}$, and weighted by the sample's sharpness, $\omega_{\text{sharpness}}$. This dual mechanism focuses the unlearning force on the causal features and adaptively gives more weight to the ``harder" samples that occupy sharper regions of the loss landscape. For the bias pathway (where $m_c=0$), the update is guided by the bias gradient, $g_{\text{bias}}$. This targeted update selectively erases the intended information from the relevant parameters, effectively preventing shortcut unlearning.

\section{Experiments}

\begin{table*}[t]
\centering
\label{tab:test_results}
\setlength{\tabcolsep}{1.7pt}
\small
\resizebox{\textwidth}{!}{%
\begin{tabular}{l c ccccc ccccc ccccc}
\toprule
\multirow{2}{*}{Method} & \multirow{2}{*}{\shortstack{w/o\\retain\\set}} & \multicolumn{5}{c}{Waterbirds} & \multicolumn{5}{c}{BAR} & \multicolumn{5}{c}{Biased NICO++} \\
\cmidrule(lr){3-7} \cmidrule(lr){8-12} \cmidrule(lr){13-17}
& & RA $\uparrow$ & FA $\downarrow$ & $\triangle_{gap}$ $\downarrow$ & WGA $\downarrow$ & MIA $\downarrow$ & RA $\uparrow$ & FA $\downarrow$ & $\triangle_{gap}$ $\downarrow$ & WGA $\downarrow$ & MIA $\downarrow$ & RA $\uparrow$ & FA $\downarrow$ & $\triangle_{gap}$ $\downarrow$ & WGA $\downarrow$ & MIA $\downarrow$ \\
\midrule
Retrain & \xmark & 100.00 & 0.00 & 0.00 & 0.00 & 10.33 & 93.84 & 0.00 & 0.00 & 0.00 & 15.88 & 77.93 & 0.00 & 0.00 & 0.00 & 24.47 \\
NegGrad & \cmark & 99.30 & 23.49 & 43.37 & 45.18 & 31.51 & 90.60 & 30.26 & 18.05 & 30.35 & 32.53 & 74.95 & 28.00 & 19.75 & 28.09 & 27.67 \\
Random label & \cmark & \textbf{100.00} & 9.34 & 18.67 & 18.67 & 25.47 & 90.84 & 36.44 & 26.70 & 36.58 & 37.01 & 75.56 & 25.55 & 15.52 & 25.62 & 40.15 \\
Bad Teaching & \xmark & 87.98 & 56.02 & 66.27 & 89.16 & 47.86 & 94.70 & 49.85 & 49.76 & 50.09 & 39.54 & \textbf{76.83} & 46.59 & 43.60 & 46.81 & 35.56 \\
Boundary Shrink & \cmark & 99.82 & 17.77 & 34.34 & 34.94 & 31.82 & 92.05 & 30.26 & 23.85 & 30.38 & 34.48 & 76.12 & 27.08 & 17.62 & 27.17 & 27.90 \\
Boundary Expand & \cmark & \textbf{100} & 11.44 & 22.89 & 22.89 & 27.54 & 91.08 & 34.38 & 26.74 & 34.51 & 35.74 & 75.88 & 27.07 & 22.85 & 27.19 & \textbf{23.98} \\
SALUN & \xmark & 99.21 & 15.06 & 25.30 & 27.71 & 30.42 & \textbf{92.53} & 53.96 & 40.06 & 54.16 & 42.75 & 77.49 & 50.94 & 47.62 & 51.18 & 34.05 \\
DELETE & \cmark & \textbf{100.00} & 8.73 & 17.47 & 17.47 & 25.86 & 92.53 & 34.38 & 25.40 & 34.51 & 35.98 & 76.66 & 22.95 & 18.72 & 23.04 & 27.36 \\
\textbf{CUPID (Ours)} & \cmark & \textbf{100.00} & \textbf{6.02} & \textbf{12.05} & \textbf{12.05} & \textbf{21.79} & 88.19 & \textbf{3.75} & \textbf{2.88} & \textbf{3.76} & \textbf{18.13} & 73.63 & \textbf{8.34} & \textbf{12.40} & \textbf{13.35} & 25.03 \\
\bottomrule
\end{tabular}
}
\caption{{Generalized Unlearning Performance on an Unbiased Test Set.} We compare various unlearning algorithms on the unbiased test sets, where bias-aligned and bias-conflicting samples are equally represented (50:50 ratio). This balanced distribution serves as the primary benchmark for evaluating generalized unlearning.}
\label{tab:main2}
\end{table*}

\subsection{Experimental Setup}
\textbf{Datasets.} To evaluate our method's ability to handle shortcut unlearning, we conduct experiments on three standard biased datasets: Waterbirds ~\cite{sagawa2019distributionally}, Biased Action Recognition (BAR) ~\cite{nam2020learning}, and Biased NICO++ ~\cite{zhang2023nico++}. These datasets are designed to test a model's reliance on spurious correlations: Waterbirds correlates bird type with background, BAR associates actions with specific places, and Biased NICO++ links object classes to particular contexts. For all datasets, the training set is heavily biased, with a 99.5:0.5 ratio of bias-aligned to bias-conflicting samples~\cite{nam2020learning}. In contrast, the test set is unbiased, with a 50:50 ratio, allowing for a fair evaluation of generalization.

\textbf{Evaluation Metrics.} To provide a comprehensive evaluation of unlearning performance under bias, we apply our metrics to samples from both the training and test sets. (1) Retain Accuracy (RA) measures performance on samples from the retain classes. (2) Forget Accuracy (FA) is the accuracy on samples belonging to the forgotten class, where a lower value is better. (3) $\triangle_{gap}$ measures the performance gap between the bias-aligned and bias-conflicting subsets of the forget set to assess whether unlearning was effectively achieved for both sample types. (4) Worst-Group Accuracy (WGA) measures the performance of the subgroup with the highest accuracy within the forget set to evaluate the balance of forgetting across all subgroups. For both $\triangle_{gap}$ and WGA, lower values are desirable as they indicate more robust and balanced unlearning. (5) Membership Inference Attack (MIA) is used to assess privacy leakage by training an attacker model to distinguish whether a given sample was part of the original training set.

We perform class-wise forgetting by randomly selecting single class from each dataset as the forget set. We use a ResNet-50 ~\cite{he2016deep} architecture for all experiments, with further implementation details deferred to the appendix. We compare our proposed method against a range of approximate unlearning algorithms. These include Retrain as the gold standard, NegGrad ~\cite{thudi2022unrolling}, Bad Teaching ~\cite{chundawat2023can}, Boundary Shrink/Expand ~\cite{chen2023boundary}, SALUN ~\cite{fan2023salun}, and DELETE ~\cite{zhou2025decoupled}.

\subsection{Experimental Results}

\textbf{Forgetting in a Heavily Biased Train Set.} We first evaluate unlearning performance directly on the training sets, an extreme scenario where the forget set is composed of 99.5\% bias-aligned samples (Table~\ref{tab:main}). This setting tests a method's ability to erase a deeply ingrained shortcut. The results demonstrate the clear superiority of our proposed method, CUPID. It achieves the lowest FA across all three datasets---6.91\% on Waterbirds, 7.70\% on BAR, and 7.71\% on Biased NICO++---closely approaching the theoretical gold standard of Retrain. This indicates that even in an environment dominated by spurious correlations, CUPID can effectively erase the influence of the target class. Furthermore, CUPID's robustness is highlighted by its minimal $\triangle_{gap}$ and lowest WGA, confirming that it performs balanced forgetting of both easy (shortcut-based) and hard (causal) information. Critically, CUPID achieves this superior performance without requiring access to the retain set, making it highly practical for real-world scenarios with privacy constraints.

\textbf{Generalized Unlearning on an Unbiased Test Set.} We then evaluate performance on the unbiased test set, which serves as the primary benchmark for assessing generalized unlearning (Table~\ref{tab:main2}). On this fair evaluation ground, CUPID again consistently and significantly outperforms all baseline methods, achieving the lowest FA across all datasets. This demonstrates that it most effectively erases the influence of the forgotten class in a way that generalizes beyond the biased training distribution. For instance, on BAR, CUPID achieves an FA of 3.75\%, whereas the next best method only reaches 30.26\%. The failure of existing methods is a clear manifestation of shortcut unlearning, which becomes evident when analyzing the $\triangle_{gap}$ and WGA metrics. Many baselines exhibit a large $\triangle_{gap}$ on this unbiased set (e.g., NegGrad shows a 43.37\% $\triangle_{gap}$ on Waterbirds), indicating they only forgot one group of samples and failed to generalize their forgetting. In contrast, CUPID maintains the smallest $\triangle_{gap}$ and lowest WGA across all datasets, confirming its ability to perform balanced and robust unlearning. While maintaining a high RA, CUPID is the only method that successfully mitigates the shortcut unlearning problem in a generalizable manner. Furthermore, CUPID demonstrates strong privacy protection, with MIA scores that are consistently closer to the Retrain than most other methods.

\begin{figure*}[t]
\centering
\includegraphics[width=0.90\linewidth]{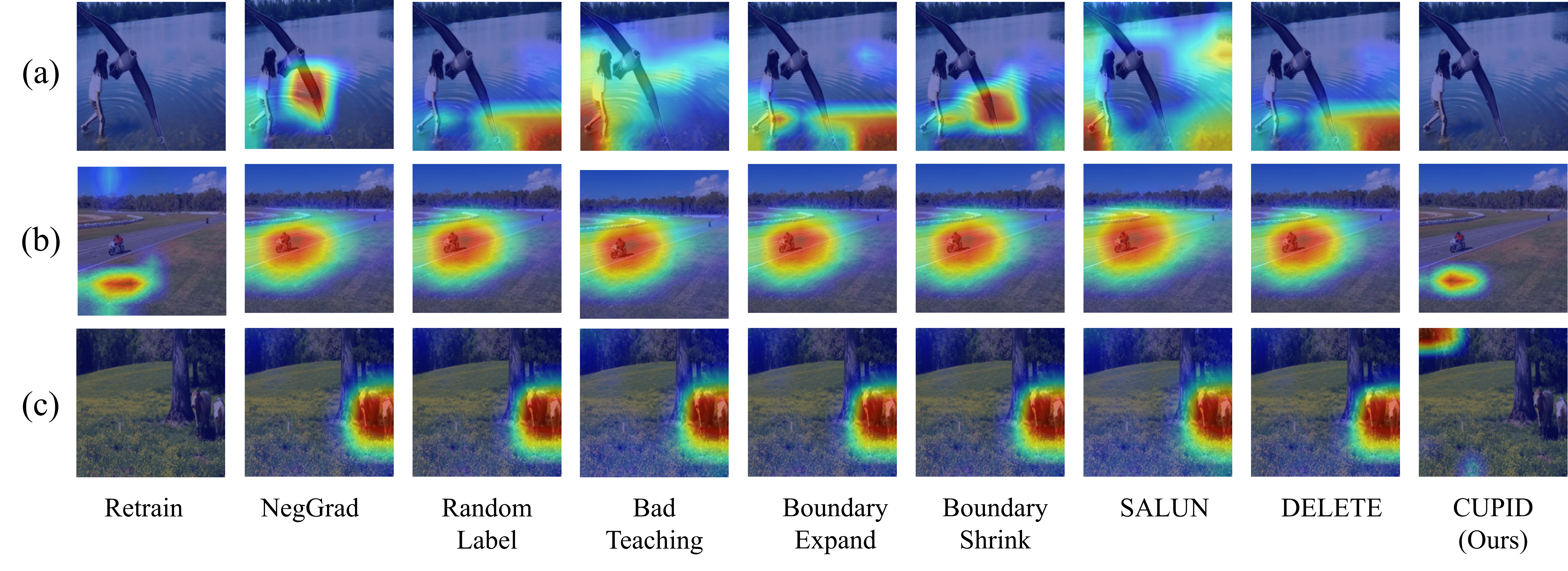}
\caption{{Qualitative Comparison using Grad-CAM.} We visualize class activation maps on three biased datasets, each designed with a specific spurious correlation: (a) Waterbirds  (b) BAR (c) Biased NICO++. The heatmaps reveal that while most existing methods continue to activate on these spurious features, our method, CUPID, successfully diverts attention from them.}
\label{fig:gradcam}
\end{figure*}

\subsection{Analysis and Ablations}

\begin{table}[t]
\centering
\small
\setlength{\tabcolsep}{6pt}
\renewcommand{\arraystretch}{1.1}
\resizebox{\columnwidth}{!}{%
\begin{tabular}{ccc|ccccc}
\toprule
\begin{tabular}[c]{@{}c@{}} (a) \end{tabular} &
\begin{tabular}[c]{@{}c@{}} (b) \end{tabular} &
\begin{tabular}[c]{@{}c@{}} (c) \end{tabular} &
RA $\uparrow$ & FA $\downarrow$ & $\triangle_{gap}$ $\downarrow$ & WGA $\downarrow$ & MIA $\downarrow$ \\
\midrule
\xmark & \xmark & \xmark & 99.81 & 34.96 & 29.17 & 36.41 & 18.44 \\
\cmark & \xmark & \xmark & 91.32 & 20.38 & 21.11 & 14.58 & 27.89 \\
\cmark & \cmark & \xmark & 99.45 & 14.56 & 15.20 & 12.78 & 19.56 \\
\cmark & \cmark & \cmark & \textbf{99.94} & \textbf{6.91} & \textbf{7.27} & \textbf{7.27} & \textbf{18.19} \\
\bottomrule
\end{tabular}
}
\caption{{Ablation Study of CUPID's Components.} We ablate our three proposed components: (a) Sharpness-Aware Partitioning, (b) Causal Pathway Identification, and (c) Targeted Pathway Update.}
\label{tab:ablation}
\end{table}

\textbf{Ablation Study.}  To verify the contribution of each component in our proposed method, we conduct an ablation study with results presented in Table~\ref{tab:ablation}. The baseline, equivalent to NegGrad ~\cite{thudi2022unrolling} (first row), shows poor performance. Introducing Sharpness-Aware Partitioning alone improves forgetting but slightly degrades RA. Subsequently adding Causal Pathway Identification not only further enhances forgetting but also restores RA to a high level, demonstrating its effectiveness in isolating updates to parameters relevant to the forget class, thereby preserving knowledge of other classes. Finally, incorporating the Targeted Pathway Update leads to a significant improvement in all forgetting-related metrics (FA, $\triangle_{gap}$, WGA), confirming its critical role in achieving effective and balanced unlearning. This step-by-step improvement demonstrates that all three components of CUPID are essential and contribute synergistically to its overall performance.

\begin{figure}[t]
  \centering
  \includegraphics[width=0.92\linewidth]{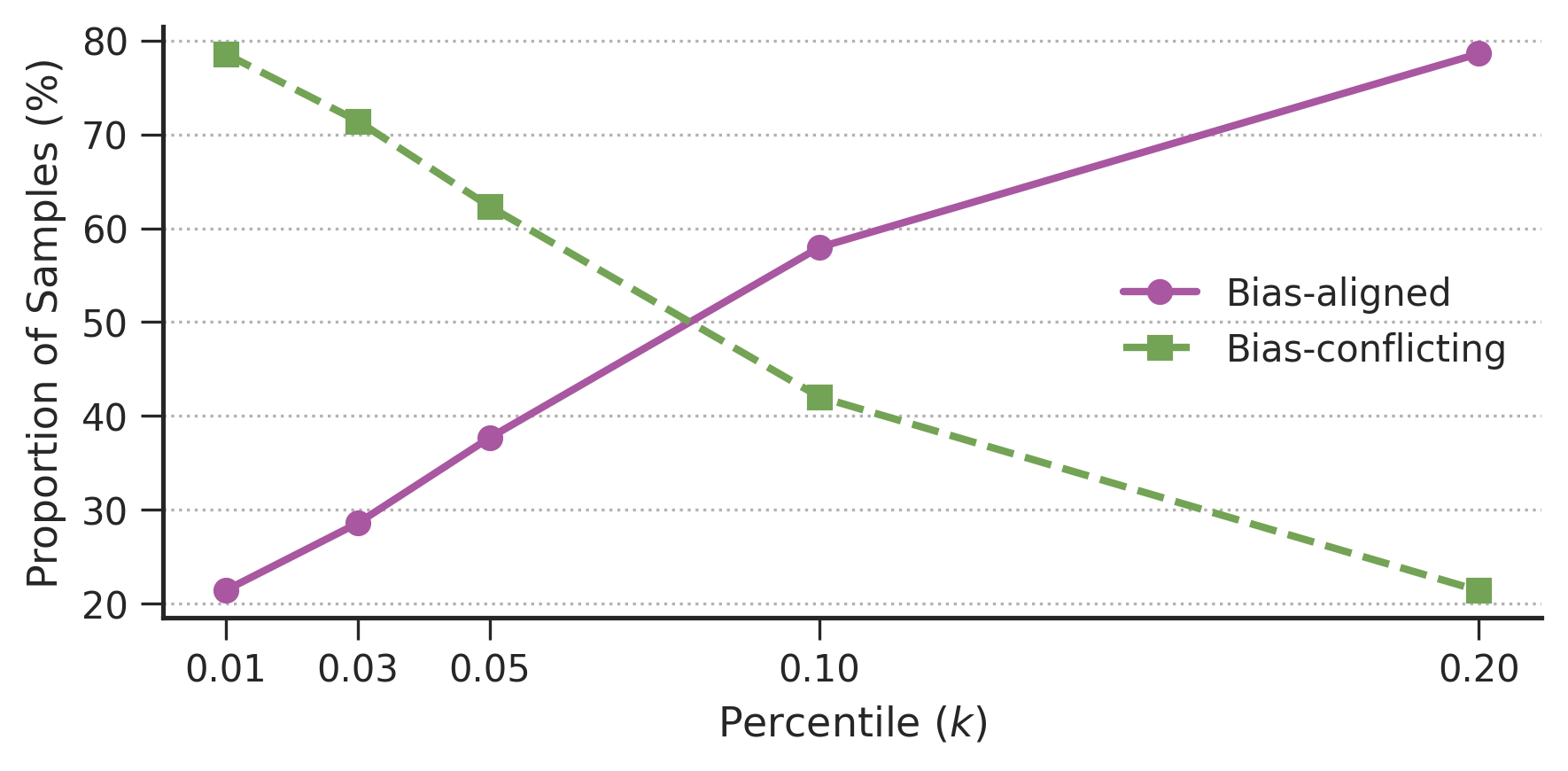}
\caption{{Composition of the Causal-Approximated Set.} The proportion of bias-aligned and bias-conflicting samples within the causal-approximated set ($D_f^{causal}$) as a function of the sharpness percentile threshold, $k$.}
  \label{fig:appendix_partition}
\end{figure}

\textbf{Analysis of Sharpness-Aware Partitioning.} To validate our Sharpness-Aware Partitioning, we analyze the resulting subset composition. Figure ~\ref{fig:appendix_partition} illustrates the proportion of true bias-conflicting and bias-aligned samples within the causal-approximated set ($D_f^{causal}$) as a function of the sharpness percentile threshold, $k$. As $k$ increases, the high sharpness threshold becomes less restrictive, leading to a predictable increase of bias-aligned samples in $D_f^{causal}$. We observe this partition is not ``pure"; however, we argue the objective is not an impossible ground-truth division, but to create functionally distinct subsets based on the model's processing. The resulting $D_f^{causal}$, while impure, is sufficiently enriched with "hard" bias-conflicting samples to compute a robust $g_{causal}$ dominated by causal features. Empirically, optimal performance occurs at $k=5\%$, rather than with a ``purer" set (smaller $k$). We hypothesize this inclusion of ``easier" bias-aligned samples regularizes $g_{causal}$, preventing noise and outlier-focus from the most difficult samples and ensuring a more stable, representative update.  

\textbf{Qualitative Comparison.} To qualitatively assess how different unlearning methods affect the model's decision-making process, we visualize the class activation maps using Grad-CAM ~\cite{selvaraju2017grad} in Figure~\ref{fig:gradcam}. The results show that existing unlearning algorithms often fail to effectively forget the target class or continue to activate regions corresponding to the bias attribute used for the original inference. In contrast, our proposed method, CUPID, not only forgets the target class but also effectively avoids activating these bias-related regions. This demonstrates that CUPID achieves a more fundamental and robust form of unlearning by correctly targeting the causal features associated with the class, rather than just manipulating the output predictions.

\section{Conclusion}
In this paper, we investigated the critical failure of machine unlearning algorithms when applied to biased models. We identified and formalized a novel phenomenon, ``shortcut unlearning," where models paradoxically struggle to forget easily-learned, bias-aligned samples and instead erase the shortcut attribute, not the intended class information. To address this, we proposed CUPID, a three-stage framework that performs a surgical unlearning intervention. By leveraging the geometric properties of the loss landscape, CUPID effectively partitions samples, identifies distinct neural pathways, and applies targeted updates to disentangle and erase causal information. Our extensive experiments on standard biased datasets demonstrate that CUPID significantly outperforms existing methods, achieving superior forgetting performance while successfully mitigating the shortcut unlearning problem. As future work, we will adapt our surgical approach to precisely target and erase abstract concepts entangled with other knowledge, mirroring the disentanglement of class and bias attributes.

\section{Acknowledgments}
This work was supported by the Institute of Information \& Communications Technology Planning \& Evaluation (IITP) grant funded by the Korea government (MSIT) [RS-2021-II211341, Artificial Intelligence Graduate School Program (Chung-Ang University)] and by the National Research Foundation of Korea (NRF) grant funded by the Korea goverment (MSIT) (RS-2025-00556246). 

\bibliography{aaai2026}

\clearpage

\section{Appendix}

\subsection{Implementation Details}

We evaluate our method on three biased datasets: Waterbirds ~\cite{sagawa2019distributionally}, which contains spurious correlations between the main class and the background; Biased Action Recognition (BAR) ~\cite{nam2020learning}, where actions are strongly correlated with specific places; and a new Biased NICO++ dataset we constructed from the original NICO++ dataset ~\cite{zhang2023nico++}. The NICO++ dataset provides rich annotations for both objects and their corresponding contexts. To construct our biased version, we first identified the most dominant context for each object class based on co-occurrence frequency. Specifically, we established strong spurious correlations by pairing objects with dominant contexts: `bird' with `dim' backgrounds, `cow' with `grass', `lizard' on `rock', `cat' near `water', `sheep' in `autumn' settings, and `dog' `outdoor'. We then curated the training set by heavily over-sampling images where the object appears in its dominant context (bias-aligned samples) and severely under-sampling images from all other contexts (bias-conflicting samples). For all datasets, the training set is constructed with a severe 99.5:0.5 ratio of bias-aligned to bias-conflicting samples to encourage shortcut learning, while the test set is balanced with a 50:50 ratio for a fair evaluation of generalization.

All experiments are conducted using PyTorch on a single NVIDIA RTX 3090 GPU. The original model is trained on each biased dataset for 10 epochs using the AdamW optimizer with an initial learning rate of $10^{-4}$, a weight decay of $10^{-3}$, and a cosine annealing scheduler. The batch size for this pre-training is 128. For the unlearning phase, all methods are run for a single epoch using the AdamW optimizer with a fixed learning rate of $10^{-5}$ and a batch size of 64. For CUPID, we set the sharpness percentile $k=5\%$ with a perturbation size $\eta$ of $10^{-3}$, and the causal pathway masking ratio $\tau_p=50\%$. 

\subsection{More Ablation Studies}

\begin{table}[h!]
\centering
\small
\setlength{\tabcolsep}{5pt}
\renewcommand{\arraystretch}{1.1}
\begin{tabular}{cc|ccccc}
\toprule
$g_{proj}$ & $g_{bias}$ & RA $\uparrow$ & FA $\downarrow$ & $\triangle_{gap}$ $\downarrow$ & WGA $\downarrow$ & MIA $\downarrow$ \\
\midrule
\xmark & \xmark & 99.74 & 18.67 & 36.14 & 36.74 & 29.86 \\
\cmark & \xmark & 99.92 & 12.77 & 12.77 & 25.24 & 30.31 \\
\xmark & \cmark & 99.91 & 17.73 & 30.21 & 31.45 & 27.21 \\
\cmark & \cmark & \textbf{100.00} & \textbf{6.02} & \textbf{12.06} & \textbf{12.05} & \textbf{21.79} \\
\bottomrule
\end{tabular}
\caption{\textbf{Synergistic Effect of Disentangled Gradients.} We analyze unlearning performance by selectively applying the causal ($g_{proj}$) and bias ($g_{bias}$) gradient components. The results demonstrate that applying both gradients to their respective pathways is critical for effective unlearning, outperforming the application of either component alone.}
\label{tab:ablation_gradient}
\end{table}

\noindent \textbf{Ablation Study on Disentangled Gradient Update.} To validate the efficacy of our distinct gradient design for the Targeted Pathway Update, we conduct an ablation study analyzing the individual and combined effects of applying the projected causal gradient, $g_{proj}$, and the orthogonal bias gradient, $g_{bias}$, to their respective neural pathways, as detailed in the Table ~\ref{tab:ablation_gradient}. The results unequivocally demonstrate the necessity of applying distinct gradients to their corresponding pathways. The baseline, which uses the standard gradient($g_f$) on the forget set, exhibits high FA and $\triangle_{gap}$ because its direction is dominated by shortcut features. Applying only the bias gradient $g_{bias}$ shows almost no improvement, as it exclusively targets the shortcut representation while leaving the core class knowledge in the causal pathway intact. Conversely, applying only the causal gradient yields a significant improvement in FA, confirming that targeting the causal pathway with a purified gradient is a crucial step. Applying both gradient components achieves the best performance across all metrics by executing a two-pronged surgical strike. It routes the purified causal gradient ($g_{causal}$) to the causal pathway to erase the class-specific knowledge, while simultaneously routing the orthogonal bias gradient ($g_{bias}$) to the bias pathway to neutralize the influence of the shortcut. This synergistic update is critical, ensuring that the class is forgotten comprehensively, which in turn prevents shortcut unlearning and leads to the most effective and balanced unlearning outcome.

\begin{table}[h!]
\centering
\small
\setlength{\tabcolsep}{5pt}
\renewcommand{\arraystretch}{1.1}
\begin{tabular}{l|ccccc}
\toprule
Method & RA $\uparrow$ & FA $\downarrow$ & $\triangle_{gap}$ $\downarrow$ & WGA $\downarrow$ & MIA $\downarrow$ \\
\midrule
w/o $\omega_{sharpness}$ & 99.91 & 17.77 & 35.54 & 35.54 & 30.81 \\
w/ $\omega_{sharpness}$ & \textbf{100.00} & \textbf{6.02} & \textbf{12.06} & \textbf{12.05} & \textbf{21.79} \\
\bottomrule
\end{tabular}
\caption{\textbf{Importance of Adaptive Reweighting.} Comparison of unlearning performance with and without using sharpness ($\omega_{sharpness}$) as an adaptive weight. The results demonstrate that this fine-grained modulation is critical for effective forgetting, complementing the initial coarse-grained partitioning.}
\label{tab:ablation_sharpness}
\end{table}

\noindent \textbf{Ablation Study on Sharpness-Aware Reweighting.} Our framework utilizes loss landscape sharpness $\omega_{sharpness}$, in two distinct stages: first for partitioning samples and second for reweighting the update. The ablation study presented in Table ~\ref{tab:ablation_sharpness} directly tests the necessity of this second stage. The results confirm that the reweighting scheme is a critical element for achieving optimal performance. Removing it, while still using sharpness for the initial partitioning, results in substantially less effective forgetting. This outcome justifies the dual role of sharpness. The initial partitioning acts as a coarse-grained filter to isolate the ``hard" causal-approximated set. However, this set is not homogenous; the degree of difficulty and informativeness for unlearning varies significantly among these samples. The reweighting then serves as a fine-grained modulator within this set. By multiplying the update by the sharpness value, our method dynamically applies a stronger unlearning force to the samples located in the sharpest, most sensitive regions of the loss landscape. 

\subsection{Additional Qualitative Results} 

\begin{figure*}[t]
\centering
\includegraphics[width=1.0\textwidth]{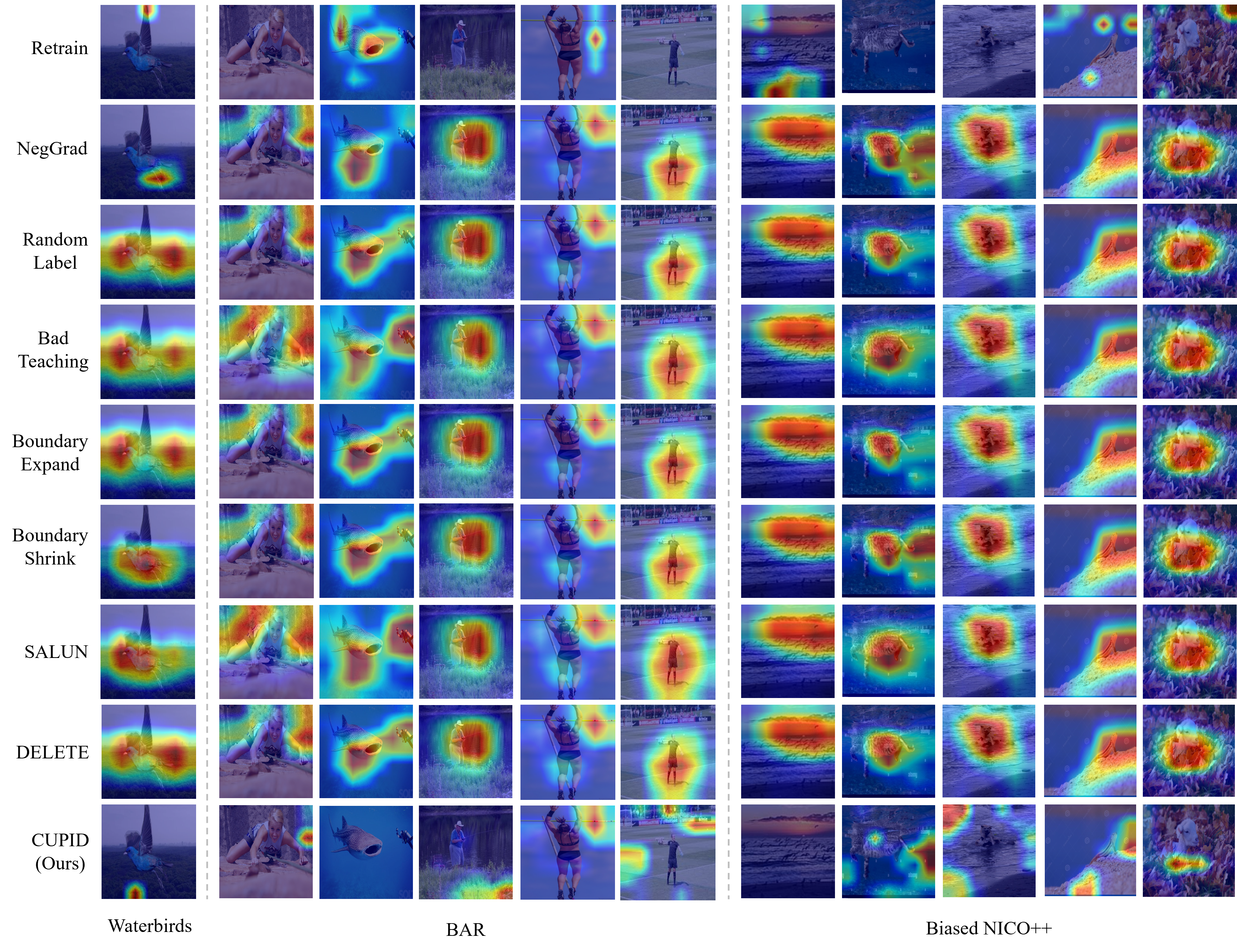}
\caption{\textbf{Qualitative Comparison of Unlearning Methods.} Grad-CAM visualizations on samples from the forget set. Most baseline methods continue to activate either the spurious shortcut or the causal object, indicating they have failed to completely forget the class. In contrast, CUPID, much like the Retrain gold standard, shows diffuse activation with no focus on either feature, demonstrating that the class concept has been successfully erased.}
\label{fig:appendix_gradcam}
\end{figure*}

We provide additional Grad-CAM ~\cite{selvaraju2017grad} visualizations in the Figure ~\ref{fig:appendix_gradcam}. 


\end{document}